\documentclass[final]{cvpr}

\usepackage{times}
\usepackage{epsfig}
\usepackage{graphicx}
\usepackage{amsmath}
\usepackage{amssymb}
\usepackage{multirow}
\usepackage{subcaption}
\usepackage[dvipsnames]{xcolor}
\usepackage{booktabs}
\usepackage{xspace}
\usepackage{interval}
\usepackage{bm}
\usepackage{mathtools}
\usepackage{footmisc}

\DeclarePairedDelimiter\floor{\lfloor}{\rfloor}
\usepackage{siunitx}
\usepackage[ruled,vlined,linesnumbered]{algorithm2e}

\SetCommentSty{mycommfont}
\SetNlSty{normalfont}{}{}

\definecolor{citecolor}{RGB}{0, 113, 188}
\usepackage[pagebackref=true,breaklinks=true,letterpaper=true,colorlinks,citecolor=citecolor, bookmarks=false]{hyperref}

\newcommand{\anet}{{\scshape ActivityNet}\xspace}
\newcommand{\fcvid}{{\scshape FCVID}\xspace}
\newcommand{\minik}{{\scshape Mini-Kinetics}\xspace}
\newcommand{\kn}{{\scshape Kinetics}\xspace}

\newcommand{\system}{{Ada3D}\xspace}
\newcommand*{\aka}{\emph{a.k.a}\@\xspace}

\def\cvprPaperID{****} 
\def\confYear{CVPR 2021}

\graphicspath{{./figures/}}

\begin{document}

\title{2D or not 2D? Adaptive 3D Convolution Selection for \\ Efficient Video Recognition}

\author{Hengduo Li$^{1}$ \qquad Zuxuan Wu$^{2}$\thanks{Corresponding author.} \qquad Abhinav Shrivastava$^{1}$ \qquad Larry S. Davis$^{1}$ \\
$^{1}$~University of Maryland \qquad $^{2}$~Fudan University \\
{\tt\small \{hdli,abhinav,lsd\}@cs.umd.edu} \quad \tt\small zxwu@fudan.edu.cn
}

\maketitle

\thispagestyle{empty}
\pagestyle{empty}

\begin{abstract}
3D convolutional networks are prevalent for video recognition. While achieving excellent recognition performance on standard benchmarks, they operate on a sequence of frames with 3D convolutions and thus are computationally demanding. 
Exploiting large variations among different videos, we introduce Ada3D, a conditional computation framework that learns  instance-specific 3D usage policies to determine frames and convolution layers to be used in a 3D network. These policies are derived with a two-head lightweight selection network conditioned on each input video clip. Then, only frames and convolutions that are selected by the selection network are used in the 3D model to generate predictions. The selection network is optimized with policy gradient methods to maximize a reward that encourages making correct predictions with limited computation. 
We conduct experiments on three video recognition benchmarks and demonstrate that our method achieves similar accuracies to state-of-the-art 3D models while requiring $20\%-50\%$ less computation across different datasets. We also show that learned policies are transferable and Ada3D is compatible to different backbones and modern clip selection approaches. Our qualitative analysis indicates that our method allocates fewer 3D convolutions and frames for ``static'' inputs, yet uses more for motion-intensive clips. 
\end{abstract}

\section{Introduction}
Videos are expected to make up a staggering 82\% of Internet traffic by 2022~\cite{videotraffic}, which demands approaches that can understand video content like actions and events accurately and efficiently.  Key to video recognition is temporal modeling to capture relationships among different frames. Towards this goal, extensive studies have been conducted with 3D convolutional networks by extending 2D convolutions over time~\cite{c3d,quovadis,r21d,channelseparated,slowfast,x3d,tpn}.  While offering excellent recognition accuracy on standard benchmarks~\cite{quovadis,sthsth,anet}, 3D models are often computationally expensive due to the costly convolution operations along the temporal axis on a large number of stacked frames. For example, at the clip-level~\footnote{Here, we use ``clip'' in a broad sense; for 2D models, a clip is a single RGB frame while for 3D models it is a stack of frames.}, a standard ResNet50~\cite{resnet} model only requires 4.1 GFLOPs (giga floating-point operations) to compute predictions for a single image, while a SlowFast network~\cite{slowfast} with the same ResNet50 backbone needs 16 times more computation (65.7 GFLOPs). Furthermore, the computational cost linearly grows with the number of clips uniformly sampled through the entire sequence for video-level prediction aggregation.

\begin{figure}[!t] \centering
    \resizebox{0.95\linewidth}{!}{\includegraphics[width=\linewidth]{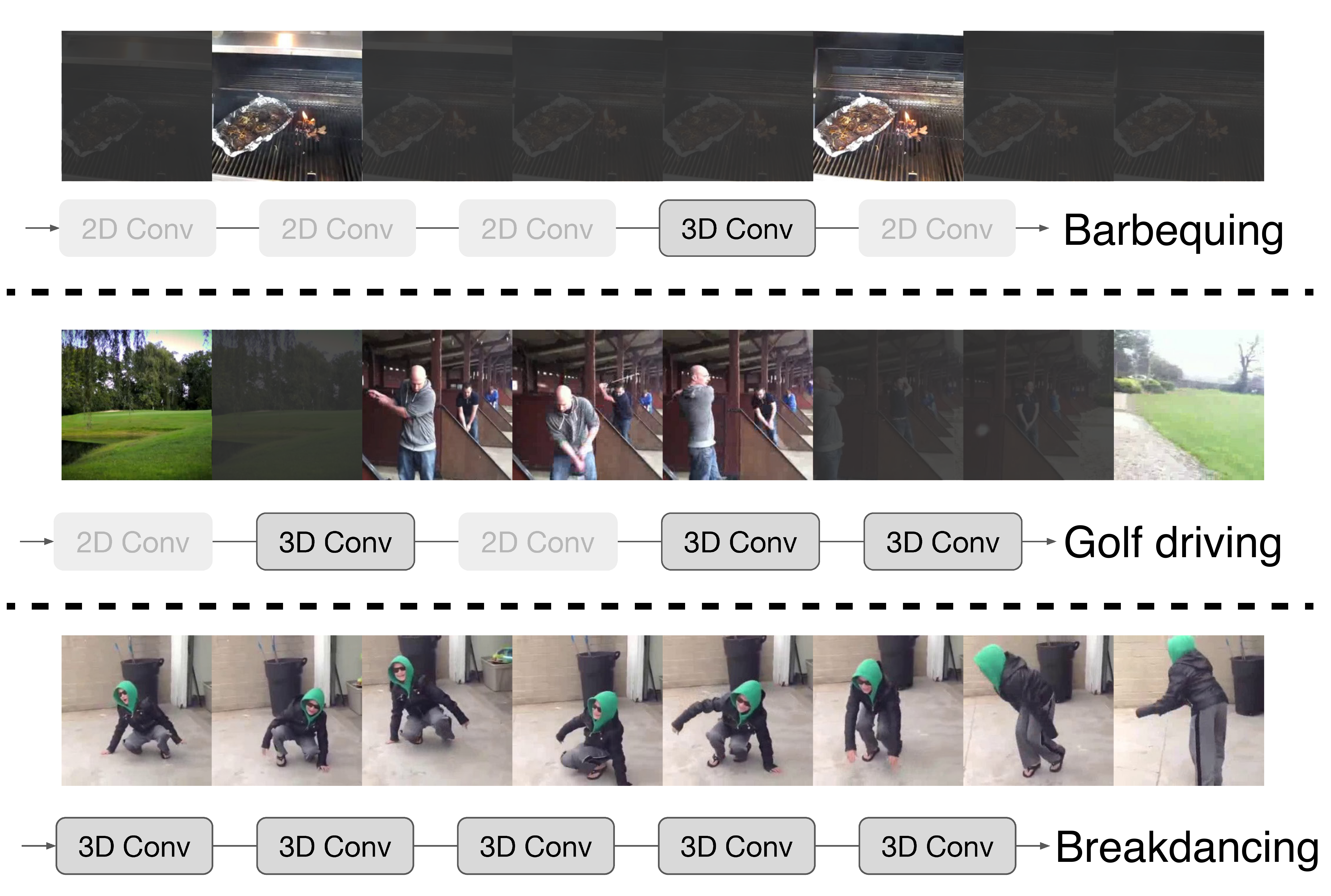}}
    \vspace{-0.13in}
    \caption{\textbf{A conceptual overview of our approach}. \system learns to adaptively keep/discard 3D convolutional layers and frames conditioned on input clips for efficient video recognition. Fewer 3D convolutions and frames are kept for clips that contain discriminative static cues and contextual information, while more are used for motion-intensive clips, in pursuit of a reduced overall computational cost without sacrificing recognition accuracy. Black mask indicates the frame is discarded.}
   \label{fig:teaser}
   \end{figure}

But are 3D convolutions really important for recognizing different types of videos? Do we really need them throughout the network? Is it necessary to perform 3D convolution on a fixed number of stacked frames for all different samples? Intuitively, 3D convolutions are critical for capturing changing patterns among inputs. However, due to large intra-class and inter-class variations, some videos are relatively more ``static'' than others, for which using a computationally expensive 3D model on redundant inputs might be unnecessary. This paper seeks to develop a computationally efficient framework for video recognition by learning how many frames to use and whether to use 3D convolutions in 3D networks. This is an orthogonal yet complementary direction to existing work on fast video recognition, which either designs lightweight 3D architectures~\cite{r21d,s3d,x3d,channelseparated} or develops clip selection schemes to use fewer clips for classification~\cite{adaframe,scsampler,listentolook,marl,dsn}.

With this in mind, we introduce \system, an end-to-end framework that learns adaptive 3D convolution usage conditioned on each input clip sample for efficient video recognition. For each clip, deriving a dynamic inference strategy entails (1) learning how many frames are used as inputs to the 3D network; (2) conditioned on these selected frames, determining how many 3D convolutional layers are activated; (3) and most importantly, making correct predictions while only using a small number of input frames and 3D convolutions. By doing so, \system allocates more computational resources to videos with complicated motion patterns while performing economical inference for ``easy static'' videos, enabling efficient video classification while maintaining reliable classification accuracy. While appealing, learning whether to keep/discard input frames and 3D convolutions is a non-trivial task, as it requires making \emph{binary} decisions that are non-differentiable. 

To this end, \system is built upon a reinforcement learning framework~\cite{policygradient}. In particular, given a video clip, \system trains a two-head selection network to produce a frame usage policy and a convolution usage policy, indicating which frames in the input stack and which 3D convolutions in the network should be kept or discarded, respectively. Then, conditioned on the derived policies, dynamic inference is performed on a pretrained 3D network with selected frames and 3D convolutions for fast recognition. The selection network is optimized with policy gradient methods~\cite{policygradient} to maximize a reward function that is carefully designed to incentivize using as few computational resources as possible while making correct predictions. We further jointly finetune the selection network with the 3D network such that the 3D model is able to adapt to the adaptive inference paradigm. It worth nothing that the selection network is designed to be lightweight so that its computational overhead is negligible.

We conduct extensive experiments to evaluate \system on ActivityNet~\cite{anet}, FCVID~\cite{fcvid}, Mini-Kinetics-200~\cite{s3d,quovadis}, and demonstrate that \system is able to save 20\% to 50\% computation on different datasets while maintaining similar recognition performance compared with baselines. We show policies learned on Mini-Kinetics-200 can be further transferred to the full Kinetics dataset~\cite{quovadis}. In addition, we show the approach is compatible with different 3D models and it is also complementary to other clip-level selection methods~\cite{scsampler,adaframe,marl,listentolook,dsn}. We also demonstrate qualitatively that our method learns to allocate fewer 3D convolutions and frames for clips that are relatively more static, while applying more computation to motion-intensive clips.

\section{Related Work} \label{sec:relatedwork}

\noindent\textbf{Deep neural networks for video recognition.} Existing work typically designs video recognition architectures by equipping state-of-the-art 2D models with the ability for temporal modeling, and can be roughly categorized into two directions. In particular, the first applies 2D models on a per-frame basis and then model temporal relationships across frames by aggregating features along the temporal axis with operations such as pooling~\cite{tsn,twostream,chris2016}, recurrent networks~\cite{recurrentdonahue,recurrentjoe,videolstm}, and using inputs with explicit temporal information such as optical flow~\cite{twostream,chris2016,tsn}. The other~\cite{quovadis,c3d,p3d,r21d,slowfast,x3d} directly transforms 2D models into 3D models with 3D convolutions applied on stacked RGB frames (clips). While achieving state-of-the-art performance on various benchmarks~\cite{quovadis,anet,sthsth}, 3D models are computationally expensive, limiting their deployment in real-world applications with limited resources. Our work aims to reduce the computational cost of 3D models by learning instance-specific 3D policies using fewer frames and 3D convolutions in a 3D model conditioned on inputs while making correct predictions at the same time.

\begin{figure*}[!t] \centering
    \resizebox{0.95\linewidth}{!}{\includegraphics[width=\linewidth]{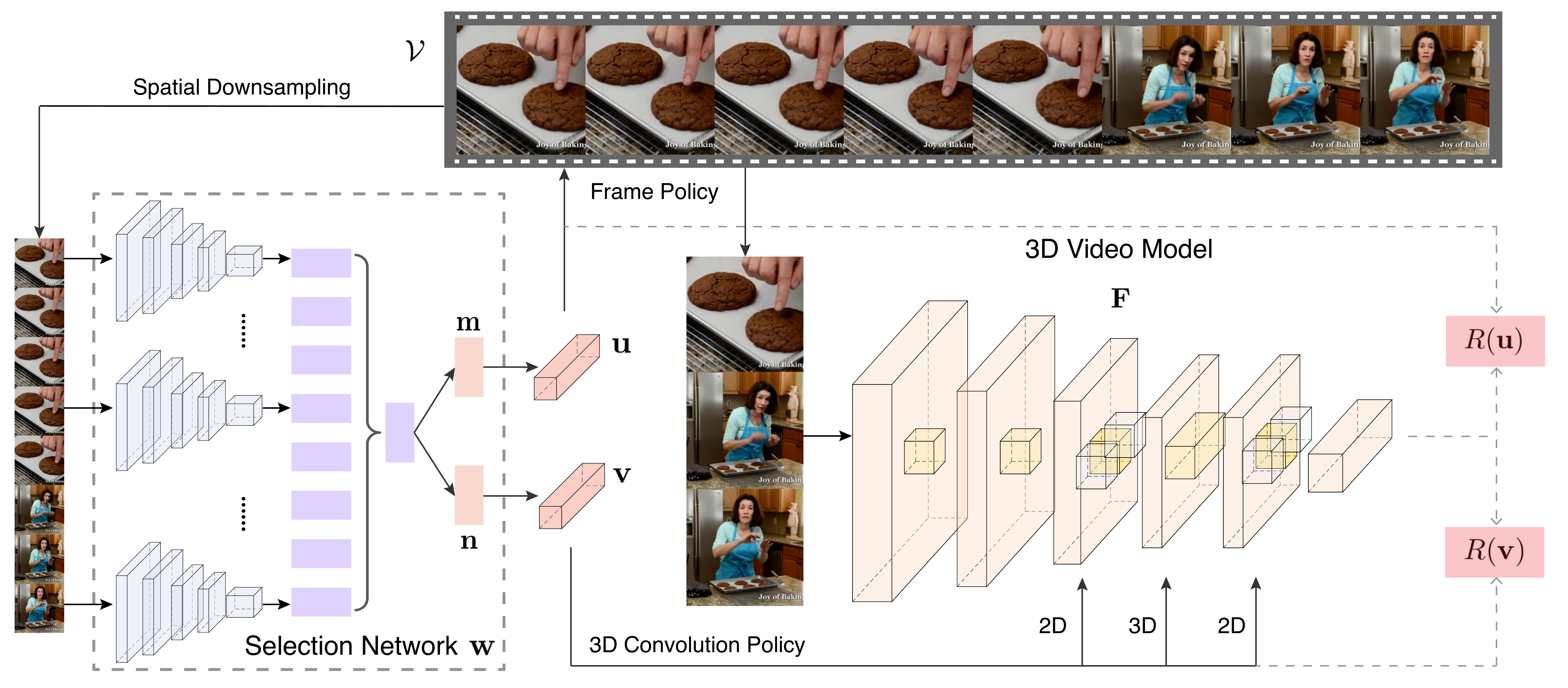}}
    \vspace{-0.15in}
    \caption{\textbf{An overview of our approach.} Given an input clip, the selection network produces features for each frame in the clip, which are further aggregated uniformly to derive a frame usage policy and a convolution usage policy simultaneously. These policies activate a subset of frames and 3D convolutions in the 3D network for inference. Then, conditioned on the prediction, two rewards are computed to evaluate the frame and convolution policy, respectively. See texts for more details.}
   \label{fig:approach}
   \end{figure*}

\vspace{0.05in}
\noindent\textbf{Efficient video recognition.} Extensive studies have been conducted on designing efficient network architectures for video recognition~\cite{eco,multifiber,r21d,x3d,trn,tsm,channelseparated}. Recent advances in efficient 2D ConvNets, \eg group convolution~\cite{mobilenets,mobilenetv2}, have been explored in 3D models~\cite{multifiber,r21d, channelseparated}. In addition, some lightweight temporal aggregation operations are introduced to speed up inference such as a relational module in TRN~\cite{trn} and a shift module in TSM~\cite{tsm}. More recently, X3D~\cite{x3d} expands a tiny model across several dimensions for a good efficiency/accuracy trade-off. However, all these approaches use a \emph{fixed}
input sampling scheme (\ie, number of frames and frame rate) and compute predictions with a ``one-size-fits-all'' model for all inputs clips, regardless of the large  temporal variations among them. In contrast, we learn dynamic frame usage policies and convolution usage policies conditioned on input clips, in pursuit of computational efficiency without sacrificing accuracy. It is worth pointing out that our method is model-agnostic, and can be used in tandem with these efficient networks. 

\vspace{0.05in}
\noindent\textbf{Adaptive computation.} Many adaptive computation (\aka, conditional computation) methods have been developed in the image domain, achieving reduced computation by dynamically selecting channels~\cite{channelgated,runtime}, skipping layers~\cite{blockdrop,figurnov,skipnet,andreasadaptive}, performing early exiting with auxiliary structures~\cite{huanggaoimproved,multiscale_densenet,icmladaptive,huanggaoresolution}, adaptively switching input resolutions~\cite{autofocus,whenandwhere,huanggaoresolution}, \etc. There are also a few recent studies exploring adaptive computation for videos. These approaches adaptively select salient clips for faster inference with one~\cite{adaframe} or more~\cite{marl} agents to aggregate video-level predictions. Compressed video~\cite{scsampler} and audio~\cite{listentolook,scsampler} are also utilized for further improvement in clip selection. More recently, a dynamic resolution selection strategy is introduced in~\cite{arnet}. 

Our method is closely related yet orthogonal to these approaches. They focus on selecting informative clips throughout the entire sequence to achieve fast inference, aiming to improve the widely used uniform sampling baseline for video recognition. For each selected clip, the same amount of computational resource is used. In contrast, we allocate computation conditioned on the complexity of the input video clip. This can be considered as dynamic routing in a network and is complementary to those clip-selection methods (as will be shown empirically)~\cite{adaframe,listentolook,scsampler}, which are a form of routing across different time steps in videos.

\section{Approach}
\system reduces the computational cost of 3D networks by learning instance-specific 3D usage policies that encourage using fewer computational resources, in the forms of frames and 3D convolutions, while producing accurate predictions. To this end, we first revisit popular 3D networks used for temporal modeling in Sec.~\ref{sec:3d}, and then elaborate different components of \system in Sec.~\ref{sec:ada3d}

\subsection{3D Networks for Video Recognition}
\label{sec:3d}
Operating on stacked RGB frames, 3D video models typically extend state-of-the-art 2D networks by replacing a number of 2D convolutions with 3D convolutions for temporal modeling over time. Formally, taking as inputs an input clip $\mathcal{V}$ with $T$ frames $\{v_1, v_2, ..., v_T\}$, 3D models obtain final predictions through a stack of 2D ($k_{1 \times d\times d}$) and 3D ($k_{t \times d \times d}$) convolutional layers, where $t$ denotes the temporal extent of 3D convolutional filters which is typically set to $3$ and $5$ in practice, and $d$ denotes the spatial height and width. In common instantiations of 3D video models~\cite{c3d,r21d,quovadis,s3d,slowfast,x3d}, 3D convolutions are inserted into the building blocks of 2D networks, and these 3D blocks are organized based on heuristics such as using them in early~\cite{s3d,r21d} or late~\cite{slowfast,s3d,tpn} stages of the network, if not applied in all stages~\cite{quovadis,x3d,r21d,c3d}. 
Note that state-of-the-art frameworks usually perform temporal convolutions in a non-degenerate form~\cite{slowfast,x3d}, \ie, taking in $T$ frames and outputting $T$  convolved frames.
While achieving state-of-the-art recognition performance, 3D video models are often computationally expensive since a number of costly 3D convolutions are applied on a sequence of stacked frames.

\subsection{\system: Adaptive 3D Convolution Selection}
\label{sec:ada3d}
\system learns 3D convolution usage policies conditioned on input video clips to reduce the computational cost of 3D models. We achieve this with a lightweight selection network that is trained to determine which frames to use as inputs to a pretrained 3D model and which convolution layers to activate in the network for those selected frames. 
This involves making binary decisions that are non-differentiable, and thus not applicable for supervised frameworks. Instead, we formulate learning the selection network as Markov Decision Process (MDP)~\cite{mdp}. We define the state space of the MDP as the input video clip; actions in the model involve keeping/discarding frames and 3D convolutions in 3D networks. The reward balances between recognition accuracy and computation.
The MDP is single-step: a video clip is observed, actions are taken, and a reward is computed---this can also be considered as a contextual bandit~\cite{bandit}.

More formally, given an input clip $\mathcal{V}$ of length $T$ and a 3D ResNet video classifier $\bf F$ with $K$ 3D convolution stages\footref{footnote:3dstage}, the selection network $f_p$, parameterized by $\bf w$, computes features for each frame in the input clip; these features are then aggregated as inputs to two parallel branches, outputting two vectors ${\bf m} \in \mathbb{R}^T$ and ${\bf n} \in \mathbb{R}^K$:
\begin{align}
{\bf m}, {\bf n}  & = \texttt{sigmoid}(f_{p}({\bf \mathcal{V}}; {\bf w})).
\end{align} 
Here, each entry in $\bf m$ and $\bf n$ is normalized to be in the range $[0, 1]$ with the $\texttt{sigmoid}(x) = \frac{1}{1 + \exp(-x)}$ function, indicating the likelihood of keeping the corresponding frame and 3D convolution stage~\footnote{We consider turning off an entire 3D convolution stage that contains multiple 3D convolutional layers to save more computation.\label{footnote:3dstage}}.

We then define a frame usage policy $\pi^f$ and a convolution usage policy $\pi^c$ with a $T$-dimensional and a $K$-dimensional Bernoulli distribution, respectively:
\begin{align}
\pi^f ({\bf u}\,|\, \mathcal{V}) & = \prod_{t=1}^{T} {\bf m}_t^{{\bf u}_t}(1-{\bf m}_t)^{1-{\bf u}_t} \\
\pi^c ({\bf v}\,|\, \mathcal{V}) & = \prod_{k=1}^{K} {\bf n}_k^{{\bf v}_k}(1-{\bf n}_k)^{1-{\bf v}_k}.
\label{eqn:bernoulli}
\end{align}
where $\bf u$ $\in \{0, 1\}^T$ and $\bf v$ $\in \{0, 1\}^K$ are \emph{actions} based on $\bf m$ and $\bf n$, and ${\bf u}_t = 1$ indicates the $t$-th frame in $\mathcal{V}$ is used; similarly ${\bf v}_k = 1$ means the $k$-th 3D convolution stage in the 3D model is activated. Zero entries in $\bf u$ and $\bf v$ represent inactive frames and convolutions, respectively. During training, ${\bf u}$ and ${\bf v}$ are produced by sampling from the corresponding policy, and a greedy approach is used at test time.

Given these actions, a subset $\mathcal{V}'$ of the full clip $\mathcal{V}$ is formed based on $\bf u$. Similarly, according to $\bf v$, certain 3D convolution layers are changed to 2D by taking only the center channel of its 3D convolutional filter along the temporal axis, \ie, the slicing operation $k_{t \times d \times d} [\floor*{\frac{t}{2}}, :, :]$ in PyTorch style. Then, conditioned on $\mathcal{V}'$, we run a forward pass with the 3D network where certain 3D convolutions are degraded, and a prediction is then computed. To encourage correct predictions with limited computation, we evaluate these actions with a reward function:
\begin{align}
R({\bf x}) = 
\begin{cases}
    1- \mathcal{O({\bf x})}   & \text{for correct prediction}\\
    -\gamma     & \text{else} \\
\end{cases}
\label{eqn:reward}
\end{align}
where $ \mathcal{O}({\bf x})$ represents the \emph{normalized} computational cost of the action and $\bf x \in \{\bf u, \bf v \}$. Based on Eqn.~\ref{eqn:reward}, we compute two rewards for frame actions and convolution actions respectively, encouraging using as little computation as possible when making correct predictions while penalizing incorrect predictions with a negative reward, \ie, $-\gamma$. Note that $\gamma$ also balances the speed-accuracy trade-off with different values. While we instantiate $\mathcal{O}({\bf u})$ and $\mathcal{O}({\bf v})$ as $(\frac{||{\bf u}||_0}{T})$ and $(\frac{||{\bf v}||_0}{K})^2$---the normalized usage of the number of frames and 3D convolutions---there are also other options such as FLOPs~\cite{arnet,amc}. The selection network is then optimized to maximize the expected reward:
\begin{align}
    \max_{\bf w}\,\mathcal{L} = \mathbb{E}_{{\bf u} \thicksim \pi_f, {\bf v} \thicksim \pi_c} [R({\bf u}) + R({\bf v})].
\end{align}

We use policy gradient methods~\cite{policygradient} to learn the parameters $\bf w$ for the selection network and the expected gradient can be derived as:
\begin{align}
\nabla_{{\bf w}}\mathcal{L}  =  \mathbb{E}\,[ & R({\bf u})\nabla_{{\bf w}}\text{log}~\pi^f ({\bf u}\,|\,{\mathcal{V}})  \nonumber\\
\, + \, &   R({\bf v})\nabla_{{\bf w}}\text{log}~\pi^c({\bf v}\,|\,{\mathcal{V}})].
\label{eqn:gradient}
\end{align}
Eqn.~\ref{eqn:gradient} can be estimated with many samples at a time, and thus we use samples in mini-batches to compute the expected gradient and then Eqn.~\ref{eqn:gradient}  is approximated by:
\begin{align}
    \nabla_{{\bf w}}\mathcal{L}  \approx  \frac{1}{B}\sum_{i=1}^{B}[ & R({\bf u}_i)\nabla_{{\bf w}}\text{log}~\pi^f ({\bf u}_i\,|\,{\mathcal{V}_i})  \nonumber\\
    \, + \, &   R({\bf v}_i)\nabla_{{\bf w}}\text{log}~\pi^c({\bf v}_i\,|\,{\mathcal{V}_i})],
    \label{eqn:gradient_empirical}
    \end{align}
where $B$ is the total number of samples in the mini-batch. The gradient is then propagated back to train the policy network with SGD. We further reduce variance by adding a baseline function to the reward~\cite{policygradient}.

\begin{algorithm}[!t]
\DontPrintSemicolon
\SetAlgoLined
\KwIn{An input video clip $\mathcal{V}$, the number of epochs of for training the selection network $E_1$, the number of epochs of joint fine-tuning $E_2$}
Obtain a pretrained video classifier $\bf F$ \\
Randomly initialize selection network ${\bf w}$ \\
\For{$e\gets 0 \:to\: E_{1}$}{
    $\bf m, n$ = \texttt{sigmoid}($f_p(\mathcal{V};{\bf w})$) \\
    $\bf u, v \sim \pi_{\bf w} ({\bf u| \mathcal{V}}), \pi_{\bf w} ({\bf v| \mathcal{V}})$ \tcp*{Eqn.~\ref{eqn:bernoulli}}
    $p = {\bf F}(\mathcal{V}|{\bf u, v})$ \tcp*{Apply actions on $\bf F$ and forward} 
    $R = R({\bf u}) + R({\bf v})$ \tcp*{Eqn.~\ref{eqn:reward}}
    ${\bf w} = {\bf w} - \nabla{{\bf w}} \mathcal{L}$ \tcp*{Eqn.~\ref{eqn:gradient}}
    }
\For{$e\gets 0 \:to\: E_{2}$}{
    Repeat Line $4$-$7$ \\
    ${\bf w} = {\bf w} - \nabla{{\bf w}} \mathcal{L}$ \tcp*{Eqn.~\ref{eqn:gradient}}
    ${\bm \theta} = {\bf w} - \nabla{{\bm \theta}} \mathcal{L}_{cls} $ \tcp*{Eqn.~\ref{eqn:jointft}}
}
\caption{Training algorithm of our approach.}
\label{alg}
\end{algorithm}

So far we have only trained the selection network while keeping the pretrained video model fixed. The selection network is able to learn decent policies that use fewer frames and 3D convolutions while maintaining prediction accuracies. However, input distributions to the 3D model are no longer the same as those used to train the original network, where all frames and 3D convolutions are used. As a result, the 3D model is not equipped with the ability to deal with inputs with varying number of frames and 3D convolutions that are adaptively turned on/off. To remedy this, we further jointly fine-tune the 3D model with the selection network such that it is able to accustomed to such adaptive inference paradigm. The objective function then becomes:
\begin{align}
& \min_{\bf w, \bm \theta} \: -\sum_{j=1}{{\bf y}^j \log({\bf F}(\mathcal{V}; {\bm \theta})^j)} - \mathcal{L}({\bf w})
\label{eqn:jointft}
\end{align}
where $\bm \theta$ denotes the weights of the 3D network $\bf F$ and the first term is the cross-entropy loss for an input clip $\mathcal{V}$ with one-hot label $\bf y$ for classification training. 
Algorithm~\ref{alg} summarizes algorithm of \system.

\begin{table*}[!t]
\centering
\setlength{\tabcolsep}{0pt} 
\begin{tabular*}{\linewidth}{@{\extracolsep{\fill}\quad}lc
    *4{S[table-format=2.1]}
    c
    *4{S[table-format=2.1]}
    c
    *4{S[table-format=2.1]}
    c@{}}
\toprule
 & & \multicolumn{4}{c}{\textbf{\fcvid}} && \multicolumn{4}{c}{\textbf{\anet}} && \multicolumn{4}{c}{\textbf{\minik}}\\ 

 \cmidrule{3-6} \cmidrule{8-11} \cmidrule{13-16}
 && {mAP} & {GFLOPs} & {\#\,3D} & {\#\,Frame} && {mAP} & {GFLOPs} & {\#\,3D} & {\#\,Frame} && {Acc} & {GFLOPs} & {\#\,3D} & {\#\,Frame} \\
 \midrule
\multicolumn{1}{l}{\bf 8-frame per clip} \\

Upper && 82.1 & 58.6 & 5.0 & 8.0 && 82.6 & 58.6 & 5.0 & 8.0 && 79.0 & 58.6 & 5.0 & 8.0 \\
Random && 78.1 & 36.1 & 2.2 & 5.8 && 79.2 & 42.9 & 3.0 & 6.6 && 74.0 & 42.2 & 2.0 & 6.9 \\
Random FT && 80.7 & 36.1 & 2.2 & 5.8 && 81.1 & 42.9 & 3.0 & 6.6 && 77.4 & 41.5 & 1.9 & 6.8 \\
Ours && 81.9 & 35.6 & 2.2 & 5.7 && 82.6 & 42.2 & 3.1 & 6.6 && 78.9 & 42.4 & 1.9 & 6.9 \\

\midrule
\multicolumn{1}{l}{\bf 16-frame per clip} \\
Upper && 84.4 & 117.3 & 5.0 & 16.0 && 84.4 & 117.3 & 5.0 & 16.0 && 79.6 & 117.3 & 5.0 & 16.0 \\
Random && 79.2 & 63.2 & 2.1 & 10.3 && 80.4 & 73.3 & 3.0 & 11.2 && 75.2 & 75.8 & 2.9 & 11.8 & \\
Random FT && 82.0 & 65.3 & 2.1 & 10.6 && 82.8 & 71.3 & 3.0 & 11.1 && 78.2 & 78.0 & 2.9 & 12.0 \\
Ours && 84.3 & 66.6 & 2.1 & 10.7 && 84.0 & 70.1 & 3.0 & 11.1 && 79.2 & 73.8 & 2.9 & 11.8 \\

\bottomrule
\end{tabular*}
\vspace{-0.05in}
\caption{\textbf{Recognition performance and computational cost of our method \vs baselines.} Two input settings are experimented, \ie 8-frame setting (\textbf{Top}) and 16-frame setting (\textbf{Bottom}). \#\,3D and \#\,Frame denote the number of 3D convolutions and frames usage per input clip respectively, averaged over the entire test set. See texts for more details.}
\label{table:main}
\end{table*}

\section{Experiments}

\subsection{Experimental Setup} \label{sec:exp_setup}
\noindent\textbf{Datasets and evaluation metrics.} We evaluate our approach on three video recognition datasets: ActivityNet (\anet)~\cite{anet}, Fudan-Columbia Video Datasets (\fcvid)~\cite{fcvid} and Mini-Kinetics-200 (\minik)~\cite{s3d}. \anet contains around $20K$ Youtube videos of $200$ action classes, with an average duration of $117$ seconds. We use the latest version $1.3$ and its official split with $10,024$ training videos, $4,926$ validation videos and $5,044$ testing videos. We report results on the validation set as the labels of testing videos are not publicly available. \fcvid consists of $91,223$ Youtube videos belonging to 239  categories, with an average duration of $167$ seconds. The official split is adopted with a training set of $45,611$ videos and a testing set of $45,612$ videos. \minik is a publicly released subset of \kn~\cite{quovadis} initially introduced in~\cite{s3d}, consisting of $200$ classes with the most training samples in Kinetics; $400$ and $25$ videos are sampled from each action class for training and validation, forming a training set with $80,000$ videos and a validation set with $5,000$ videos. Here we use the identical samples as~\cite{s3d}. To demonstrate the transferability of the selection network, we experiment with the Kinetics full set, which contains 240K training videos and 20K validation videos.

Following official instructions, we report mean average precision (mAP) on \anet and \fcvid. For \minik and \kn, we report Top-1 accuracy.

\vspace{0.05in}
\noindent\textbf{Network architectures.} We use an I3D~\cite{quovadis} with a backbone of ResNet-50~\cite{resnet} as the 3D video model if not mentioned otherwise, due to its popularity and competitive recognition performance across various benchmarks~\cite{quovadis,sthsth,anet}. Our implementation follows~\cite{slowfast_repo}, where 3D convolutions are factorized spatially and temporally in a similar way as R(2+1)-D~\cite{r21d}, which is already a more efficient architecture than original I3D. In addition, we also experiment with the Slowonly model introduced in~\cite{slowfast} to demonstrate the compatibility of our approach with more recent networks.

We use a lightweight architecture for the selection network with negligible computational overhead. Specifically, we use MobileNetV2~\cite{mobilenetv2} as the backbone of the selection network. The inputs to the network are downsampled to $112 \times 112$ per frame, and it only requires $0.08$ GFLOPs to compute features for each frame.

\vspace{0.05in}
\noindent\textbf{Implementation details.}
All 3D networks are fine-tuned from models provided by~\cite{slowfast_repo}, which are pre-trained on Kinetics. We fine-tune 3D models for 40 epochs on \fcvid and \anet and $20$ epochs on \minik, with a cosine learning rate schedule starting at $0.01$ and a batch size of $64$. The MobileNetV2 backbone of the selection network is also pre-trained on these datasets with the same schedules to speed up convergence. We first fix the pretrained 3D models and train the selection network for $40$ epochs with a learning rate of $0.0001$ and a batch size of $256$. Finally, the whole pipeline is jointly fine-tuned for $60$ epochs with the same learning rate described above. SGD with momentum $0.9$ is used for optimization. We use $8$ GPUs for all experiments. 

Regarding network inputs during training, we follow~\cite{slowfast,x3d} by randomly sampling a clip with $8$/$16$ frames using a temporal stride of $8$ (sampling rate) from a given video. For the spatial domain, $224 \times 224$ pixels are randomly cropped from the sampled clip during training. For inference, we follow the common practice~\cite{slowfast,x3d,nonlocal} and uniformly sample $10$ clips with a spatial size of $256 \times 256$ from a testing video. Video-level prediction is obtained by averaging the clip-level predictions.

\subsection{Main Results}
We compare our proposed method with various baselines under different input settings ($8$/$16$ frames per clip) and report results in Table~\ref{table:main}. The baselines we use include:
\begin{itemize} \itemsep0.1em
    \item \emph{Random}: Based on the frame usage and convolution usage produced by \system, we generate random policies that use a similar amount of computational resources compared to \system.    
    \item \emph{Random FT}: The 3D model is further jointly fine-tuned with the random policies.
    \item \emph{Upper}: The original pretrained 3D model with all 3D convolutions and all frames used, which can be viewed as a performance ``upperbound'' of our method.
\end{itemize}

As shown in Table~\ref{table:main}, under the 8-frame input setting, \system obtains an mAP (accuracy for \minik) of 81.9$\%$, 82.6$\%$ and 78.9$\%$, requiring an average of 35.6, 42.2 and 42.4 GFLOPs per clip on \fcvid, \anet, \minik respectively. 
\system achieves comparable recognition performance but brings 40$\%$, 28$\%$ and 27$\%$ computational savings. This confirms that \system is able to learn effective 3D convolution and frame usage policies by saving computational resources and preserving accuracies at the same time across different datasets. Similar patterns are also observed under the 16-frame input setting for all three datasets.

Using similar computational resources, \system improves the \emph{Random} baseline by $3.5\%$ to $5\%$ mAP/accuracy on three datasets. \system also outperforms \emph{Random FT} by $1\%$ to $2.5\%$. These results verify that \system produces adaptive polices and allocates computational resources on a per-input basis to maintain recognition performance. It is worth noting that there are slightly differences in computational savings on different datasets. This results from the fact that video categories in these datasets are different. For example, FCVID contains some classes of static objects and scenes like ``bridge'' and ``temple'', and thus we observe more computational savings than \anet and \minik, which are more activity-focused; on \minik, where categories are motion-intensive, more computational resources are needed compared to \fcvid and \anet.

\begin{figure}[!b] \centering
   \resizebox{0.9\linewidth}{!}{\includegraphics[width=\linewidth]{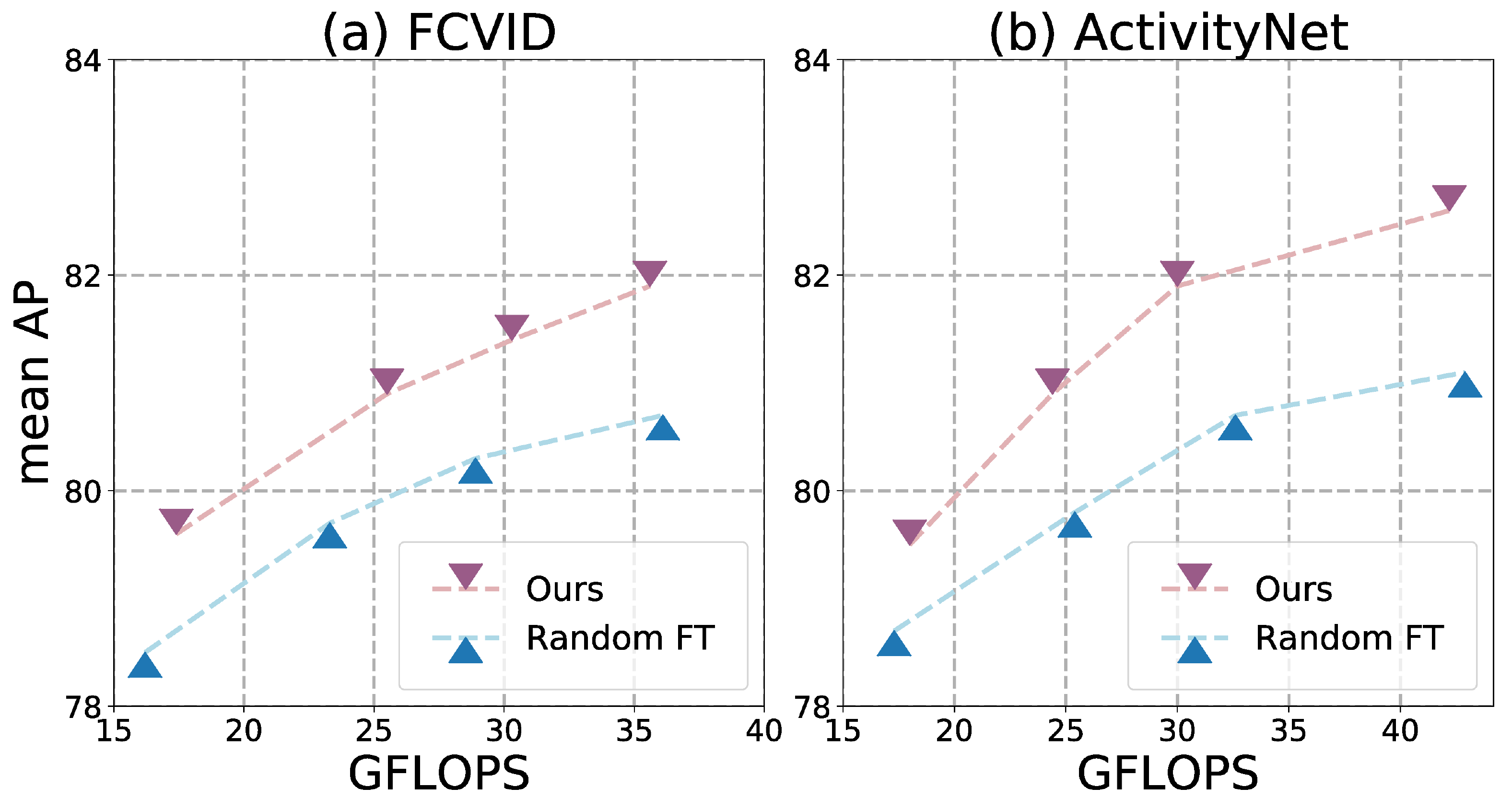}}
   \vspace{-0.12in}
   \caption{Recognition performance under different computational budgets controlled by $\gamma$.}
   \label{fig:ablation}
\end{figure}

\vspace{0.05in}
\noindent\textbf{Recognition with varying computational budgets.} As discussed in Section~\ref{sec:ada3d}, the choice of $\gamma$ in Eqn.~\ref{eqn:reward} adjusts the amount of penalty on policies that produce incorrect predictions, and thus it controls the speed/accuracy trade-off. Here we report recognition accuracies of \system under different computational budgets. As demonstrated in Fig.~\ref{fig:ablation}, our method is able to cover a wide range of speed/accuracy trade-offs and consistently outperforms \emph{Random FT} with different computational budgets. For example, on \anet, \system obtains an mAP of 82.6$\%$, 81.9$\%$ and 80.9$\%$ with an average of 42.2, 30.1 and 24.4 GFLOPs per clip respectively, while \emph{Random FT} obtains 81.1$\%$, 80.7$\%$ and 79.8$\%$ with 42.9, 32.6 and 25.4 GFLOPS per clip on average. Same patterns are also observed on \fcvid.

\begin{table}[!t] \centering
    \resizebox{1.0\linewidth}{!}{
    \begin{tabular}{*{10}c}
    \toprule
      && \multicolumn{2}{c}{\emph{Van} $\rightarrow$ \emph{Ada}}  && \multicolumn{2}{c}{\emph{Van} $\rightarrow$ \emph{Ada}} && \multicolumn{2}{c}{\emph{Van} $\rightarrow$ \emph{Ada}} \\
    \cmidrule{1-1} \cmidrule{3-4} \cmidrule{6-7} \cmidrule{9-10}
    \#\,Clip && \multicolumn{2}{c}{3.0 $\rightarrow$ 2.9} && \multicolumn{2}{c}{5.0 $\rightarrow$ 4.2} && \multicolumn{2}{c}{10.0 $\rightarrow$ 7.4} \\
    mAP && \multicolumn{2}{c}{77.8 $\rightarrow$ 78.1} && \multicolumn{2}{c}{80.3 $\rightarrow$ 80.5} && \multicolumn{2}{c}{81.9 $\rightarrow$ 82.0} \\
    \bottomrule
    \end{tabular}}
    \vspace{-0.1in}
    \caption{\textbf{Extension to clip-level selection.} Combining \system with AdaFrame~\cite{adaframe} offers computational savings for video-level aggregation. \#\,Clip denotes number of clips used per testing video; \emph{Van} (Vanilla) and \emph{Ada} denote our method without and with AdaFrame, respectively.}
    \label{table:adaframe}
    \end{table}

\vspace{0.05in}
\noindent\textbf{Extension to clip selection.} As mentioned in Sec.~\ref{sec:relatedwork}, our method is orthogonal and thus could be complementary to the line of clip selection methods~\cite{marl,adaframe,scsampler,listentolook} for efficient video recognition. We validate our hypothesis by combining our method with AdaFrame~\cite{adaframe}. Specifically, we use \system as the backbone of AdaFrame to dynamically allocate computational resources conditioned on each input clip, as opposed to the original AdaFrame that uses the same amount of computation with a \emph{fixed} backbone for all clips. Following~\cite{adaframe}, we train three variants of AdaFrame which operates on 3, 5, and 10 clips for different computational budgets. As demonstrated in Table~\ref{table:adaframe}, extending our approach with adaptive clip selection further decreases the computational cost while producing comparable performance with the \emph{Upper}. For example, it reduces the number of clips sampled from each testing video from $10$ to $7.4$ and obtains an mAP of 82.0$\%$ that is on par with \emph{Upper} (82.1$\%$). Additionally, we believe our method is also complementary to other clip selection methods leveraging multi-modal inputs such as audio~\cite{scsampler,listentolook}, as well as adaptive spatial resolution modulating methods~\cite{arnet,whenandwhere,huanggaoresolution}.

\begin{table}[!h]  \centering
    \resizebox{0.8\linewidth}{!}{\begin{tabular}{*{6}c}
    \toprule
    Method && Acc & GFLOPs & \#\,3D & \#\,Frame \\
    \cmidrule{1-1} \cmidrule{3-6}
    Upper && 73.1 & 58.6 & 5.0 & 8.0 \\
    Ours && 72.8 & 43.7 & 2.3 & 6.9 \\
    \bottomrule
    \end{tabular}}
    \vspace{-0.1in}
    \caption{\textbf{Transferring learned policies.} We fine-tune a Kinetics pretrained model on Kinetics full training set, with policies learned on Mini-Kinetics, and evaluate on Kinetics validation set.}
    \label{table:kinetics}
    \end{table}

\noindent\textbf{Transferring learned policies.} We now analyze whether the policies learned by our method can be transferred to novel action categories. To this end, we take the selection network trained on \minik and fine-tune a pretrained I3D model with a ResNet-50 as its backbone on full Kinetics. We keep the weights of the selection network fixed during fine-tuning. Details of training and testing are the same as joint fine-tuning as described in Sec.~\ref{sec:exp_setup}.
As shown in Table~\ref{table:kinetics}, policies learned on \minik can reduce the overall computational cost of the fine-tuned video model by $25\%$ on Kinetics with negligible difference in recognition accuracy compared to the \emph{Upper} baseline, indicating that our method learns strategies that are transferable to unseen classes and videos. It is worth noting that the I3D baseline we use obtains superior recognition performance on Kinetics that is higher than~\cite{quovadis,s3d} and competitive compared to results reported in~\cite{nonlocal} using 32 frames per input clip.

\begin{figure*}[!t] \centering
    \resizebox{1.0\linewidth}{!}{\includegraphics[width=\linewidth]{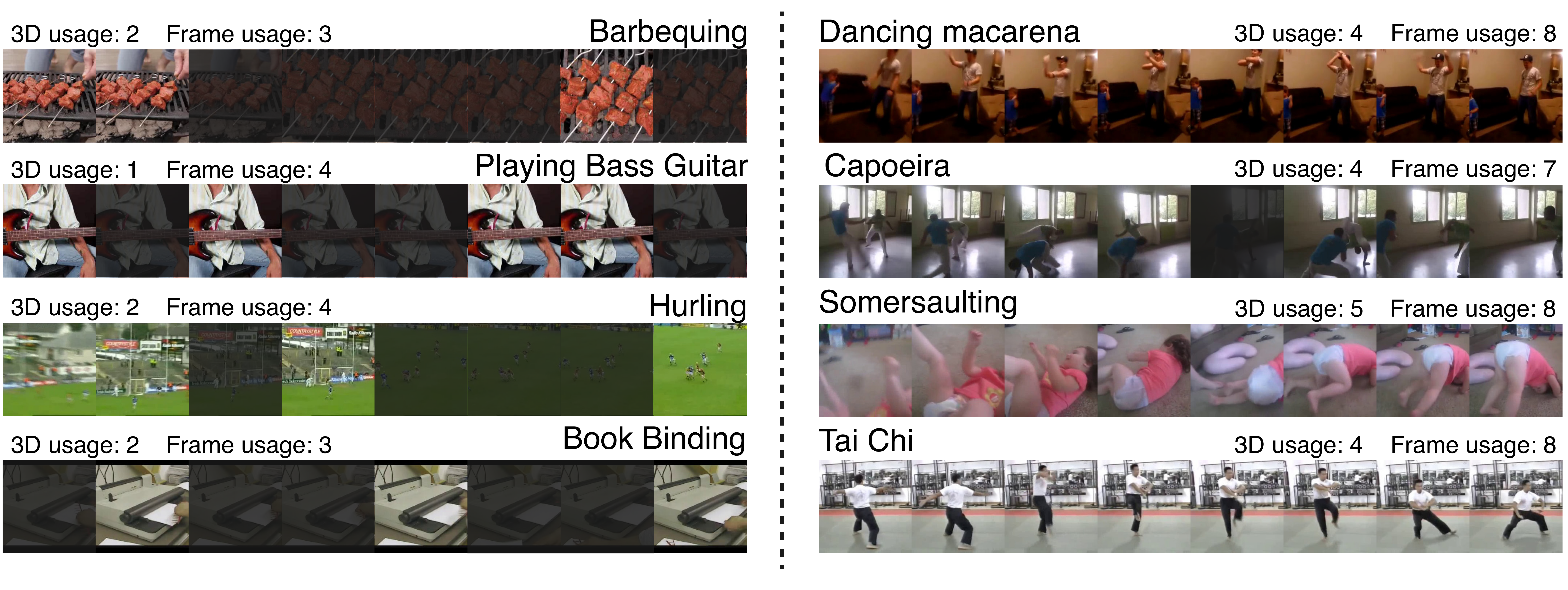}}
    \vspace{-0.35in}
    \caption{\textbf{Qualitative results.} Black mask indicates the frame is discarded. \textbf{Left:} Fewer 3D convolutions and frames are used for action classes and instances that are more ``static'', \ie containing discriminative static cue and contextual information. \textbf{Right:} For motion-intensive instances, more computation is allocated for probing finer temporal information.}
   \label{fig:quals}
   \end{figure*}
\vspace{0.05in}
\noindent\textbf{Compatibility with different 3D architectures.} Next, we evaluate the compatibility of our approach with different 3D networks. We use a more efficient 3D network architecture recently introduced in~\cite{slowfast} termed as Slowonly and evaluate our approach. In particular, it only uses 3D convolutions in the 4-th and 5-th stage of a ResNet50, resulting in competitive recognition performance with less computational cost. As shown in Table~\ref{table:slow}, our method still obtains $20\%$ to $40\%$ savings in GFLOPs with similar recognition performance, indicating \system is compatible with different 3D models. Our method by design is model-agnostic, for which we believe it could be complementary to recent work on designing efficient 3D models such as X3D~\cite{x3d} as well. 

\begin{table}[h]  \centering
    \resizebox{0.8\linewidth}{!}{\begin{tabular}{*{6}c}
    \toprule
    Method && mAP & GFLOPs & \#\,3D & \#\,Frame \\
    \cmidrule{1-1} \cmidrule{3-6}
    Upper && 82.6 & 54.5 & 2.0 & 8.0 \\
    Ours && 82.4 & 42.1 & 1.3 & 6.6 \\
    \cmidrule{1-1} \cmidrule{3-6}
    Upper && 83.5 & 109.1 & 2.0 & 16.0 \\
    Ours && 83.4 & 61.8 & 1.4 & 9.5 \\
    \bottomrule
    \end{tabular}}
    \vspace{-0.1in}
    \caption{Results on FCVID~\cite{fcvid} using Slowonly~\cite{slowfast} architecture as 3D model. \textbf{Top}: 8-frame input setting. \textbf{Bottom}: 16-frame input setting.}
    \label{table:slow}
    \end{table}

\vspace{0.05in}
\noindent\textbf{Qualitative analysis.} In addition to the quantitative results presented above, we also qualitatively analyze our method. In particular, we observe that our method produces policies with fewer 3D convolutions and frames for input clips that are more ``static'', while uses more for motion-intensive instances. As shown in Fig.~\ref{fig:quals}, a smaller number of 3D convolutions and frames are applied on clips with discriminative static cue. For instance, the presence of ``bass'' and ``book binder'' for class ``playing bass guitar'' and ``book binding'' suffice to produce correct predictions, and the scene of a ``court'' serves as a strong contextual signal for ``hurling''. On the other hand, for motion-intensive action classes and instances, especially those related to human movement such as ``breakdancing'', ``somersaulting'' and ``Tai Chi'', more computational resources are allocated by our method to capture finer temporal relationships among frames.

\subsection{Discussion}
\begin{table}[t!]  \centering
    \resizebox{!}{1.7cm}{\begin{tabular}{*{8}c}
    \toprule
    \multicolumn{2}{c}{} && \multicolumn{2}{c}{\textbf{\fcvid}} && \multicolumn{2}{c}{\textbf{\anet}} \\
    \cmidrule{1-2} \cmidrule{4-5} \cmidrule{7-8}
    Tr & FT && mAP & GFLOPs && mAP & GFLOPs \\
    \cmidrule{1-2} \cmidrule{4-5} \cmidrule{7-8} 
    \multicolumn{2}{c}{Upper} && 78.1 & 58.6 && 76.4 & 58.6 \\
    \cmidrule{1-2} \cmidrule{4-5} \cmidrule{7-8} 
    & && 72.3 & 36.1 && 71.1 & 42.9 \\
     \checkmark & && 75.9 & 34.7 && 74.3 & 38.1 \\
     & \checkmark && 76.5 & 34.9 && 75.1 & 37.6 \\
    \checkmark & \checkmark && 77.8 & 35.6 && 76.1 & 42.2 \\
    \bottomrule
    \end{tabular}}
    \vspace{-0.12in}
    \caption{Ablation on the effectiveness of two training stages.}
    \label{table:ablation_components1}
\end{table}

\noindent\textbf{Impact of joint finetuning.} Recall that we first train the seletion network with the 3D model fixed and then jointly fine-tune both of them. Here we analyze the performance of our method without the first selection network training stage (Tr) or the joint fine-tuning stage (FT). For faster evaluation, we uniformly sample 3 clips from each test video. Results are shown in Table~\ref{table:ablation_components1}.

As can be seen, joint fine-tuning is crucial to further improve the recognition performance (75.9 \vs 77.8). This indicates that fine-tuning the video model together with learned policies indeed helps the 3D model to adapt to the adaptive inference paradigm brought by the selection network. It is worth noting that skipping the first training stage (\ie, directly training the selection network with the 3D model jointly) leads to a lower recognition performance (76.5 \vs 77.8). We posit the reason is that adding another objective (the classification loss) while training the selection network from random initialization further increases the instability of network learning under such a reinforcement learning setting; and thus the selection network converges to sub-optimal policies.

\begin{table}[h!]  \centering
    \resizebox{!}{1.7cm}{\begin{tabular}{*{8}c}
    \toprule
    \multicolumn{2}{c}{} && \multicolumn{2}{c}{\textbf{\fcvid}} && \multicolumn{2}{c}{\textbf{\anet}} \\
    \cmidrule{1-2} \cmidrule{4-5} \cmidrule{7-8}
    3D & Frame && mAP & GFLOPs && mAP & GFLOPs \\
    \cmidrule{1-2} \cmidrule{4-5} \cmidrule{7-8} 
    \multicolumn{2}{c}{Upper} && 78.1 & 58.6 && 76.4 & 58.6 \\
    \cmidrule{1-2} \cmidrule{4-5} \cmidrule{7-8} 
    & && 75.5 & 35.6 && 74.3 & 42.3 \\
     \checkmark & && 76.8 & 35.3 && 75.3 & 41.1 \\
     & \checkmark && 76.3 & 35.5 && 74.8 & 43.5 \\
    \checkmark & \checkmark && 77.8 & 35.6 && 76.1 & 42.2 \\
    \bottomrule
    \end{tabular}}
    \vspace{-0.12in}
    \caption{Ablation on the usefulness of 3D convolution usage and frame usage policies.}
    \label{table:ablation_components2}
\end{table}

\vspace{-0.1in}
\noindent\textbf{Contributions of convolution and frame usage policies.} To demonstrate the effectiveness of 3D convolution usage and frame usage policies learned by the two-head selection network, we conduct experiments to analyze contributions of the two components. In particular, we replace each/both components with randomly generated policies similar to \emph{Random FT}. Here we use 3-clip testing as well. As shown in Table~\ref{table:ablation_components2}, applying either 3D or frame usage policy improves recognition performance under the same computational budget, while using both achieves the best performance with $1\%$ improvement over the single-component settings, indicating the double-head architecture can learn to produce policies cooperatively.

\section{Conclusion}
We presented \system, a framework that learns to derive adaptive 3D convolution and frame usage policies---determining which 3D convolutions in a pretrained 3D video model and which frames in the input clip to use on a per-input basis---for efficient video recognition. In particular, a two-head selection network is trained with policy gradient methods to produce these policies, reducing overall computational cost while maintaining recognition performance. Extensive experimental results on three large-scale video recognition datasets indicate that \system achieves 20$\%$-50$\%$ computational savings on state-of-the-art 3D video models while achieving similar accuracies. We further demonstrate \system is compatible with different backbones of 3D model and other clip selection methods, and qualitatively show that more computational resource is allocated on motion-intensive instances but less on static ones by \system.

\textbf{Acknowledgement} \small This work is supported by IARPA via Department of Interior/Interior Business Center (DOI/IBC) contract number D17PC00345.

{\small
\bibliographystyle{ieee_fullname}
\bibliography{reference}
}

\end{document}